\documentclass[twoside,leqno,twocolumn]{article}

\usepackage[letterpaper]{geometry}

\usepackage{ltexpprt}
\usepackage{hyperref}

\usepackage{multirow}
\usepackage{graphicx}
\usepackage{amsmath}
\usepackage{amssymb}
\usepackage{subfigure}

\begin{document}

\newcommand\relatedversion{}

\title{\Large Reinforcement Learning Guided Multi-Objective Exam Paper Generation \relatedversion}

\author{Yuhu Shang \thanks{These authors contributed to the work equally and should be regarded as co-first authors.} \thanks{Tianjin University of Science \& Technology. shangyuhu6902@mail.tust.edu.cn, zhxkun@tust.edu.cn, renyimeng@mail.tust.edu.cn, liangkun@tust.edu.cn.} 
\and Xuexiong Luo \footnotemark[1] \thanks{Macquarie University. xuexiong.luo@hdr.mq.edu.au.} 
\and Lihong Wang \thanks{National Computer Network Emergency Response Technical Team Coordination Center of China. wlh@isc.org.cn.
}
\and Hao Peng \thanks{Beihang University. penghao@buaa.edu.cn.}
\and Xiankun Zhang \footnotemark[2] \thanks{Corresponding author.}
\and Yimeng Ren \footnotemark[2]
\and Kun Liang \footnotemark[2]
}
\date{}

\maketitle

\fancyfoot[R]{\scriptsize{Copyright \textcopyright\ 2023 by SIAM\\
Unauthorized reproduction of this article is prohibited}}





\begin{abstract} \small\baselineskip=9pt {To reduce the repetitive and complex work of instructors, exam paper generation (EPG) technique has become a salient topic in the intelligent education field, which targets at generating high-quality exam paper automatically according to instructor-specified assessment criteria. The current advances utilize the ability of heuristic algorithms to optimize several well-known objective constraints, such as difficulty degree, number of questions, etc., for producing optimal solutions. However, in real scenarios, considering other equally relevant objectives (e.g., distribution of exam scores, skill coverage) is extremely important. Besides, how to develop an automatic multi-objective solution that finds an optimal subset of questions from a huge search space of large-sized question datasets and thus composes a high-quality exam paper is urgent but non-trivial. To this end, we skillfully design a reinforcement learning guided Multi-Objective Exam Paper Generation framework, termed MOEPG, to simultaneously optimize three exam domain-specific objectives including difficulty degree, distribution of exam scores, and skill coverage. Specifically, to accurately measure the skill proficiency of the examinee group, we first employ deep knowledge tracing to model the interaction information between examinees and response logs. We then design the flexible Exam Q-Network, a function approximator, which automatically selects the appropriate question to update the exam paper composition process. Later, MOEPG divides the decision space into multiple subspaces to better guide the updated direction of the exam paper. Through extensive experiments on two real-world datasets, we demonstrate that MOEPG is feasible in addressing the multiple dilemmas of exam paper generation scenario\footnote{\href{https://github.com/researcher-tiger/MOEPG}{https://github.com/researcher-tiger/MOEPG}}}.
\end{abstract}
\section{Introduction}\label{sec:1}
Examination is an essential means to distinguish examinees’ abilities and select talents, which plays an important role in computer-assisted education and adaptive learning \cite{bib1}. Manually browsing and composing exam papers by instructors is inefficient because of the exponential number of feasible combinations of questions. As such, the advances in computer technology have far-ranging consequences in practical applications for large-scale web-based examination.

\par{
A promising approach for supporting web-based examination is exam paper generation (EPG) \cite{bib2,bib3}. Recently, heuristic techniques such as Genetic Algorithm \cite{bib4, bib5}, Particle Swarm Optimization \cite{bib7, bib8} have become one of the most pervasive tools in EPG domain. They focus on optimizing multiple assessment criteria, such as difficulty degree, the number of questions, to constitute an exam paper. Although these methods are straightforward to implement, they still face the following problems: (1) they tend to optimize several well-known objectives, especially focus more on the difficulty degree of the generated exam paper while ignoring other equally relevant objectives (e.g., distribution of exam scores, skill coverage). In actual tasks, an elegant exam paper should simultaneously optimize all the objectives and trade-off among conflicting objectives; (2) they rely on experts to label the difficulty degree of the question, which may leads to some efforts on EPG task somehow are deviating from the examinee group’s cognitive level; and (3) they follow a straightforward but inefficiency global sampling strategy to adjust the exam paper.}


\par{In light of the above, it motivates us to think about three key technical issues. First, the increase of the dimension of optimization objectives may lead to the poor performance of the heuristic techniques in solving models with multiple conflict objectives \cite{bib9}. We note that deep reinforcement learning \cite{p1, p2, p3} has been applied to combinatorial optimization in recent years and has achieved convincing results. However, the DRL technique is difficult to use because the amount of computation increases when the action space is large. Second, existing methods usually mine linear interactions of examinee learning process by manually designed function (e.g., logistic function). Nevertheless, due to the complexity of human knowledge acquisition, oversimplification may cause inaccurate diagnostic results. Therefore, an advanced deep learning model is needed that can predict each examinee’s exam score when modeling the functional relationship between the learning records and the examinees’ skill proficiency. Third, the traditional global sampling strategy faces the issue of the excessive decision space due to the huge number of questions in the question set. A more effective way to divide the decision space into multiple subspaces is required for EPG to guide the updated direction of exam paper.}

\par{Inspired by the above motivation and intuition, we discuss scenarios wherein the EPG needs to jointly optimize for multiple objectives (i.e., difficulty degree, distribution of exam scores, and skill coverage) and propose a new EPG model called reinforcement learning guided Multi-Objective Exam Paper Generation framework (MOEPG) as a solution. More concretely, the optimization process of EPG can be regarded as a Markov Decision Process (MDP) where the agent successively selects a sequence of operations, i.e., the right actions based on different versions of the exam paper thus optimize the predefined multiple conflicting objectives. Then, in order to simulate the exam scores of examinee group, we adopt Deep Knowledge Tracing (DKT) \cite{bib10} to obtain the examinee's skill proficiency by analyzing the history of the feedback on questions. Thus, we can easily calculate the difficulty degree of the exam paper and the distribution of the student group’s exam scores. Meanwhile, we design a question set partition mechanism that cluster related or similar questions into a same group to better guide the update direction of exam paper. The present work proposes the following key contributions: }

\par{1. \textbf{(Uniform Scheme)} MOEPG holds the flexible exam Q-Network, a function approximator, to optimize multiple conflicting objectives simultaneously, maximizing the total quality of the generated exam paper.}
\par{2. \textbf{(Automatic Proficiency Assessment)} Considering the inaccuracy of manually labeling question difficulty, we integrate the advanced deep knowledge tracing into the MOEPG framework to estimate the skill proficiency of the examinee group.}

3. \textbf{(Effective Sampling Strategy)} MOEPG divides the decision space into multiple subspaces to help the agent realize partition sampling, alleviating the inherent issue of large action space on reinforcement learning based methods.

\section{Related work}
\subsection{Exam Paper Generation.}
Roughly, exam paper generation approaches mainly lie in three categories. The first category centers around random methods such as random selection algorithm \cite{bib11} and shuffling algorithm \cite{bib12}. The random method randomly extracts questions from a large-sized question pool, and the generated exam paper may have an arbitrary difficulty degree. The second category centers around finding questions according to the difficulty degree of the exam paper. For difficulty objective, many work tries to explore various model structures (e.g., fuzzy logic algorithm, genetic algorithm) to compose appropriate exam paper \cite{bib13,bib14,bib15}. 
The third category of research highlight the multi-objective nature of EPG task.
The work in \cite{bib6, bib16, bib17} regard a question as chromosome that constitutes an exam paper, and jointly integrate several objectives (e.g., difficulty degree, discrimination degree, exam time, etc.) as evolutionary objectives.
Nguyen et al. \cite{bib7} presents an exam paper generation approach using particle swarm optimization in which they take fifty questions and get the optimal solution for them.
The work in \cite{bib2} presents an exam paper generation method under the nested combination of difficulty degree, discrimination degree, and related topic using integer programming.
However, the significant drawback is that they usually consider several well-known objectives (e.g., difficulty degree, discrimination degree, etc.) while ignoring other equally relevant objectives. Besides, none of these models consider simplification of sampling strategy during the optimization process.

\subsection{Student Performance Prediction.}
Obtaining the skill mastery level of examinee group is one of the most significant requirements of EPG task. In the literature, numerous KT based attempts have been made, including Bayesian Knowledge Tracing (BKT) \cite{bib19}, and Performance Factor Analysis (PFA)\cite{bib20}. Especially the Deep Knowledge Tracing (DKT) \cite{bib10}, achieved the state-of-the-art KT task, which can work as the prefer for evaluating examinees’ mastery level of multiple skills. Nowadays, DKT is widely applied in adaptive learning, such as knowledge recommendation \cite{bib21}, and educational gaming \cite{bib22}.
 In our work, we utilize the acquired skill proficiency to predict each examinee’s exam scores. Besides, DKT can discover the latent correlation among skills during training, a task that is typically required to be pre-labeled by experts \cite{bib23}.
 \section{Preliminaries}
\emph{Definition 1: (Formalization of Learning System Entities)}. Suppose there are $|E|$ examinees, $|Q|$ questions, and $|K|$ skills. Each examinee’s learning records can be represented as a sequence $\mathcal{X}=\{ x_1, x_2, \cdots ,x_{|\mathcal{X}|} \}$, $x_t=\{q^{(t)}, y_t \}$, where $q^{(t)}$ is the question that the examinee attempts at the timestamp $t$, $ y_t =\{0,1 \} $ is the response score to question $q^{(t)}$, ${|\mathcal{X}|}$ denotes the sequence length of a certain examinee. The knowledge skill set in the system is represented by $K=[ k_1, k_2, \cdots ,k_{|K|} ]$. 

\noindent \emph{Definition 2: (Exam Paper Specification)}. The attributes of the exam paper specification $\mathcal{M}$ can be expressed as $\mathcal{M}=\{n,b,o,d\}$ where $n$ represents the number of questions specified for the exam, $b$ indicates score for each question, $o$ indicates the overall score of the exam, $d$ represents the predefined difficulty degree of the exam. The exam paper generation process aims to find a subset of questions from a question set $Q=\{ q^{(1)}, q^{(2)}, \cdots ,q^{(|Q|)} \}$ to form an exam paper with specification $\mathcal{M}$ that maximizes the predefined multiple objectives. Then the exam paper is defined as follows:
\begin{equation}\label{e.3.1}
 \begin{aligned}
   \mathcal{M}_q = \begin{bmatrix} 
			q(k_1^1) & q(k_2^1) & \dots & q(k_{|K|}^1)\\
			q(k_1^2) & q(k_2^2) & \dots & q(k_{|K|}^2) \\
			& \vdots & \ddots & \vdots \\
			q(k_1^n) & q(k_2^n) & \dots & q(k_{|K|}^n)
		\end{bmatrix}_{n \times |K|},
 \end{aligned}
\end{equation}
\par{\noindent where each element of $M_q$ is 0 or 1. If $i$-th skill is covered by question $q^{(j)}$, then $q(k_i^j)$=1, otherwise $q(k_i^j)$=0.}
\section{Methodology}
\par{The workflow of MOEPG is presented in Figure \ref{fig1}. In the subsequent sections, the details and connections among three core mechanisms are elaborated upon, in Sections \ref{sec:4.1} (Exam Score Prediction), \ref{sec:4.2} (Exam Q-Network), and \ref{sec:4.3} (Question Set Partition).}
\subsection{Mechanism 1: Exam Score Prediction.}\label{sec:4.1} 
\begin{figure*}
\centering
\includegraphics[width=\linewidth]{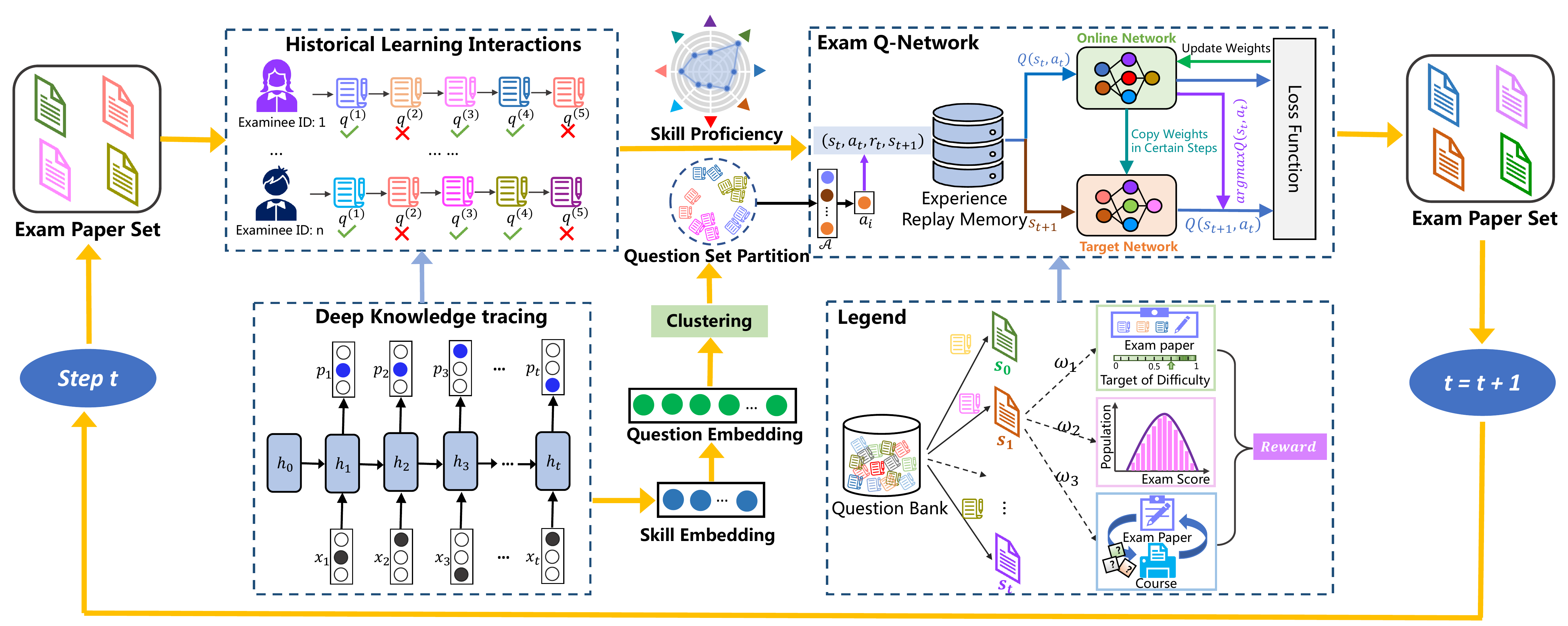}
\caption{Workflow of the proposed MOEPG. First, MOEPG predicts each examinee’s exam performance by estimating their skill proficiency. Then, MOEPG adopts Double Deep Q-Networks (DDQN) based training algorithm to achieve an acceptable tradeoff among multiple exam domain-specific objectives. Finally, MOEPG skillfully divides the decision space into multiple subspaces to guide the direction of exam paper updating. }\label{fig1}
\end{figure*}

One of the most essential parts of the MOEPG, is the prediction of the examinee exam score. 
An explicit obtain the skill proficiency of the examinee group, making the model predicting examinee exam score more effective. Therefore, we design a deep knowledge tracing (DKT) based exam score prediction mechanism, which accurately attains the student mastery levels of skills to predict the student's exam scores. As shown in the bottom left part of Figure \ref{fig1}, the input ($x_t$) of the DKT is the examinee’s past learning records, and the prediction ($p_t$) represents the probability of a specific skill being mastered. The process can be expressed as:
\begin{equation}\label{e.4.2}
\begin{aligned}
	i_{t} &= \sigma( {W_{xi}x_{t} + W_{hi}h_{t - 1} + b_{i}} ),\\
	f_{t} &= \sigma( {W_{xf}x_{t} + W_{hf}h_{t - 1} + b_{f}} ),\\
	o_{t} &= \sigma( {W_{xo}x_{t} + W_{ho}h_{t - 1} + b_{o}} ),\\
	c_{t} &= f_{t}{c}_{t - 1} + i_{t} tanh( {W_{xc}x_{t} + W_{hc}h_{t - 1} + b_{c}} ),\\
	h_{t} &= o_{t}tanh(c_{t} ),\\
	p_{t} &= \sigma({W_{s}o_{t} + b_{s}}),\\
	\end{aligned}
	\end{equation}
\par{\noindent where $i_*$, $f_*$, $c_*$, $o_*$, $h_*$ are the input gate, forget gate, output gate, cell activation vector, and hidden state respectively. $W_*$ and $b_*$ are the learned parameters.}

Then, $p(K^e)=[ p(k_1^e), p(k_2^e), \cdots,p(k_{|K|}^e)]$ stores the $e$-th examinee’s mastery of all skills, where the elements $p{(k_i^e)}\in[0,1]$ represents the predicted probability that the examinee can correctly answer the $i$-th skill. Thus, the skill proficiency of examinee group is defined as:
\begin{equation}\label{e.4.3}
\begin{aligned}
	\mathcal{P} = [p(K^1), p(K^2), \dots, p(K^{|E|})].
\end{aligned}
\end{equation}
During training, the model parameters are learned by minimizing the binary cross-entropy loss between the predicted probability $p_t$ and the true label $y_t$ as,
\begin{equation}\label{e.4.4}
 \begin{aligned}
   \mathcal{L}=-\sum_{e}^{|E|}\sum_{t=1}^{|\mathcal{X}_e|}\left(y_tlogp_t + (1-y_t)log(1-p_t)\right) ,
 \end{aligned}
\end{equation}
\noindent where ${|\mathcal{X}_e|}$ represents the length of examinee e’s question-answering sequence in the training set.

Later, we utilize the acquired skill proficiency to calculate the predicted probability $r_{e, j}$ that the $e$-th examinee would correctly answer the $j$-th question:
\begin{equation}\label{e.4.5}
 \begin{aligned}
  r_{e, j} = \prod_{i=1}^{|K|}p{(k_i^e)}^{q{(k_i^j)}}.
 \end{aligned}
\end{equation}
Generally, the final score of an exam is not known until the end of the exam. In our work, we can predict each examinee’s exam scores $r_{e}$ in a priori manner:
\begin{equation}\label{e.4.6}
 \begin{aligned}
   r_{e} = \sum_{j=1}^{n} (r_{e,i} \times b_j), 
 \end{aligned}
\end{equation}
where $b_j$ denotes the score of the $j$-th question.
\par{Then, the exam score of the examinee is determined as: $R_s=[ r_1, r_2, \cdots r_{|E|} ]$. For convenience, we define the exam score distribution of the examinee group as $P(R_s)$.}
\subsection{Mechanism 2: Exam Q-Network.}\label{sec:4.2}
\par{In the previous section, we record the exam score distribution into $P(R_s$ by deep knowledge tracing. Next, we adopt the double deep Q-Network (DDQN) \cite{bib24} to update the exam paper. As a RL problem, $<S,A,R,T>$ in the MDP are defined as:
}

$ \bullet $ State $S$: 	A state $s_t \in S$ reflects the status of the exam paper update process at time $t$, i.e., $s_t = [q_*^{(1)} \oplus q_*^{(2)} \oplus \dots \oplus q_*^{(n)}]$, where $\oplus$ is the concatenation operation, and $q_*^{(i)}$ is the feature vector of the $i$-th question in the exam paper.

$ \bullet $ Action $A$: Based on state $s_t$, taking action $a_t \in A$ is defined as replacing a question in the exam paper with a new question from the question set. 

$ \bullet $ Reward $R$: The agent receives an immediate reward $r_t$ according to the status of the $t$-th version of exam paper, where we design it with multiple objectives.

 $\bullet $ Transitions $T$: Once the agent selects a question $p^{(j)}$ from the question set, we will update $s_t$ to $s_{t+1}$ by removing one question from $s_t$ and adding the new question $p^{(j)}$, thus keeping a fixed size (i.e., $n$) of the exam paper. Specifically, we try to use a new question to replace each question in the exam paper in turn and record the subscript $h$ of the question with the highest reward value. Then, $h$-th question in the exam paper is replaced by new question $p^{(j)}$. Subsequently, $s_t$ will be updated as $s_{t+1} = [p^{(j)}\oplus q_*^{(1)} \oplus \dots \oplus q_*^{(n-1)}]$.
 
\par{We now elaborate on the flow of updating the exam paper. At step $t$, the agent acquires the state $s_t$ of the current exam paper and takes an action $a_t$ (i.e., extracting a question) from a global question set with an $\varepsilon$-greedy policy. Then, the agent receives the reward $r_{t+1}$ by simulating the examinee answering the exam paper and updating the observed state to $s_{t+1}$. Eventually, we store the experience $(s_t,a_t,r_{t+1},s_{t+1})$ into the experience replay memory.}
\par{Furthermore, $N_b$ size data is randomly put into the network, and map this information to $Q(s, *, \theta)$, where $\theta$ can be trained to help select the most appropriate question. The target network (with parameter $\theta^-$) and the online network (with parameter $\theta$) share the same structure. The parameters of the online network are synchronized to the target network in certain steps. The target Q-value $y_t^{DDQN}$ can be written as:}
\begin{equation}\label{e.4.7}
\resizebox{.98\hsize}{!}{$y_t^{DDQN} = r_{t+1} + \gamma Q(s_{t+1}, argmax_{a_{t+1}} Q(s_{t+1}, a_{t+1};\theta);\theta^- )$.}
\end{equation}
\par{The deep neural network utilizes a non-linear function approximator $\theta$ to estimate the action-value function, i.e., $Q(s_t,a_t;\theta)$.
By minimizing the mean-square loss function, we can estimate this approximator.}
\begin{equation}\label{e.4.8}
L_t(\theta_t) = \mathbb{E}_{(s_t,a_t,r_{t+1},s_{t+1})\sim Me} \left [ \left( y_t^{DDQN}-Q(s_t,a_t;\theta) \right)^2 \right ].
\end{equation}
\par{In process of exam paper update, the traditional global sampling strategy commonly faces the issue of the excessive decision space due to the huge number of questions in the question set. Besides, the questions selected randomly may not be evenly spread across the question set. To this end, we develop a partition-based sampling strategy which will be used in a partitioning fashion for locating the range of action selection.}
\subsection{Mechanism 3: Question Set Partition.}\label{sec:4.3}
\par{There has been a lot of work offering their own perspectives on how to reduce the action space \cite{bib25,bib26}. Different from these works, we divided the potentially related candidate action space into multiple subspaces and select questions in the subspace for restricted updates. Additionally, the optimization direction of exam paper gives threshold $ts$ as the control target. Specifically, once the question of a certain subset is selected by the agent, the system will calculate the corresponding skill coverage reward. A subset other than the current subset is selected as the candidate action space if the skill coverage reward for the current question in the selected subset is below the threshold $ts$. Therefore, the agent should go to other subspaces to find and match questions. Otherwise, the agent continues to find questions in the current subspace to meet the other two objectives (i.e., difficulty degree, distribution of exam score). Therefore, how to partition the question set is crucial. And the following elaborates on this process:}
\par{Step1: Train DKT model for at most 100 epochs to obtain best performances. Based on the well-trained DKT, the $i$-th skill embedding vector is represented as:}
\begin{equation}\label{e.4.9}
emb_{k_i}=emb(k_i;W_*).
\end{equation}
\par{Step2: The $j$-th question is represented by the embedding distribution of knowledge skills, i.e.,}
\begin{equation}\label{e.4.10}
emb_{q_j}= \sum_{i=1}^{|K|}emb(k_i\cdot{q{(k_i^j)}};W_*),
\end{equation} 
where $k_i\cdot{q(k_i^j)}$ is used to record the skill contained in the $j$ -th question.
\par{Step3: K-means is used to cluster the questions. Thus, the final question set $F$ is divided into $f$ subsets, each containing $p_f$ questions.}
\begin{equation}\label{e.4.11}
\begin{aligned}
&F =\{ F_{1\sim{p_{1}}}^{(1)},F_{1\sim{p_{2}}}^{(2)},\ldots,F_{1\sim{p_{{f}}}}^{(f)} \}. \\
\end{aligned}
\end{equation}
\par
\begin{figure}[htbp]
\centering
\includegraphics[width=0.8\linewidth]{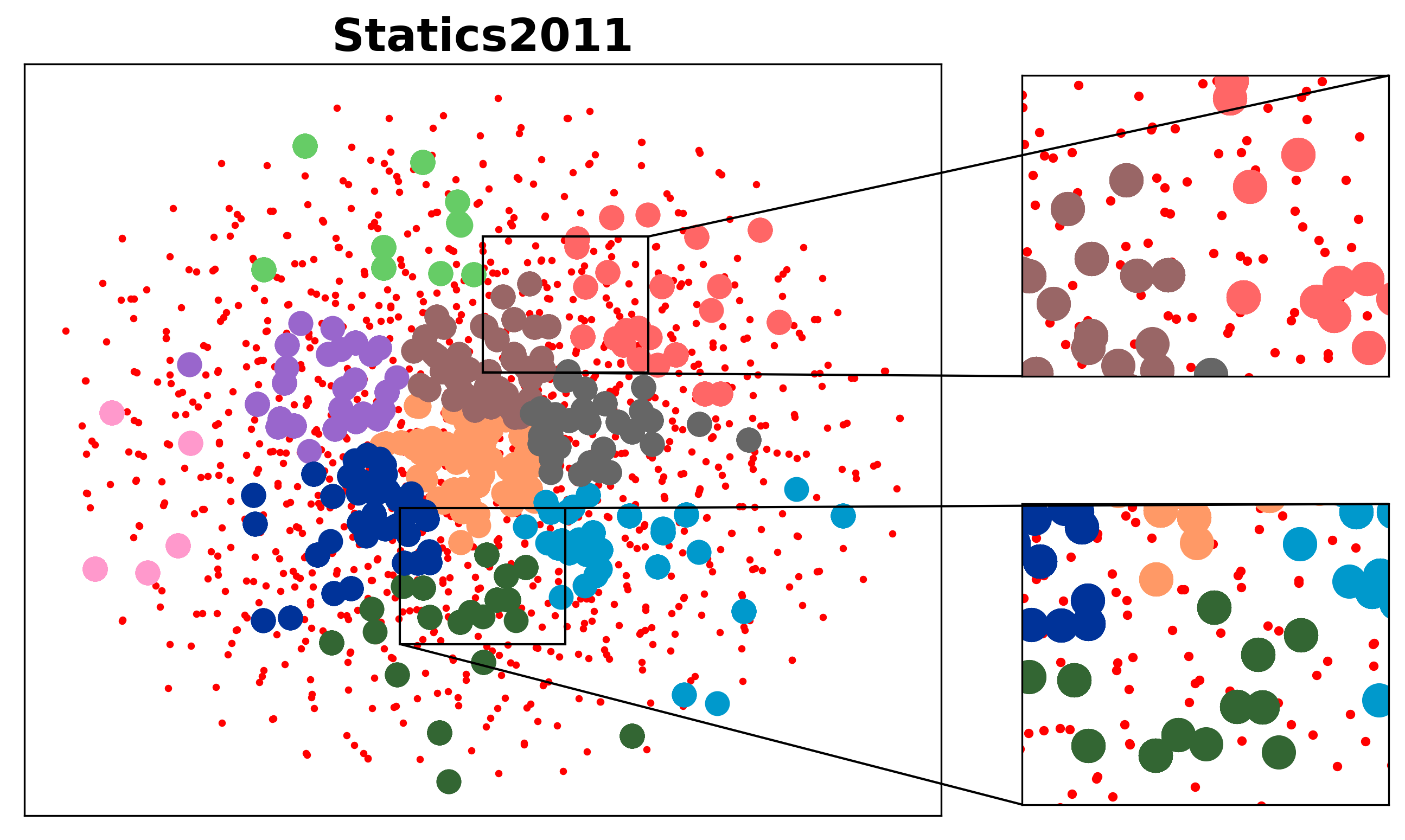}
\caption{Question relation on Statics 2011 dataset.}\label{fig2}
\end{figure}

{ We try to visualize the skill embeddings (red spots) and question embeddings (other color spots) in Statics2011 using t-SNE tool (see Figure \ref{fig2}). All questions are grouped into $f$ clusters, where the questions from the same cluster (skill) are labeled in the same color. This distribution shows that questions associated with the same skill tend to be located together, while questions belonging to different clusters are well separated. This shows the capability of MOEPG to discover the implicit relationships of questions and apply them to optimize the updating process of the exam paper.}
\vspace{-0.2cm}
\subsection{Multi-Objective Rewards.}\label{sec:4.4}
Next, we will focus on the design of reward function $r$. Three different objectives w.r.t. difficulty degree, distribution of exam score, and skill coverage are used in this paper. Then, we design our reward function in a multi-objective way.

\par{\textbf{Difficulty Degree.} The difficulty degree of the exam paper should approximate the difficulty degree requirement given by instructors. Then, $r_1$ is used to measure the distance between the average exam scores of student group $\bar{R}$ and the desired exam difficulty $\mu$.}
\begin{equation}\label{e.4.10}
r_1 = 1-\mid \mu-\bar{R} \mid.
\end{equation}
\par{\textbf{Distribution of Exam Score.} The exam score of the examinee group is influenced by multiple uncertain factors \cite{bib27}. Thus, the distribution of exam scores $P(R_s)$ should satisfy the desired normal distribution $N(\mu,\sigma^2)$ where $\mu$ denotes the average score of all examinees and $\sigma^2$ denotes the ability to distinguish academic performance between examinees. Then, $r_2$ is defined as the differences between $P(R_s)$ and $P(Z)$:}
\begin{equation}\label{e.4.11}
\begin{aligned}
r_2 &= Wasserstein(P(R_s),P(Z)) \\
&= 1 - \inf \limits_{\gamma \sim \prod(P(R_s),P(Z))} E_{(x,y) \sim \gamma}[|| {x-y} ||],\\
\end{aligned}
\end{equation}



\par{\noindent where $\prod((P(R),P(Z)))$ represents the set of all joint distributions $\gamma({x,y})$ whose marginals are respectively $ P(R)$ and $P(Z)$. Intuitively, $\gamma({x,y})$ represents how much "mass" must be transported from $x$ to $y$ in order to transform $ P(R)$ into $P(Z)$ \cite{bib32}.
}
\par{\textbf{Skill Coverage.} In practice, the proportion of each skill of an exam paper ($\bar{V}$) must satisfy the relative importance of skills in the course $(\bar{C})$ \cite{bib33}. Here, we apply a commonly used Cosine similarity to measure the similarity between $\bar{V}$ and $\bar{C}$, which is defined as,}
\begin{equation}\label{e.4.12}
r_3=Different(\overline{V}, \overline{C}) = \sum_{i=1}^{|K|}\frac{v_i\cdot c_i}{||v_i|||c_i||},
\end{equation}
\par{\noindent where $v_i$ represents the proportion of $ i $-th skill to be covered within the exam, and $ c_i $ represents the relative importance of $ i$-th skill in the course. For instance, $v_i$ can be obtained by:}
\begin{equation}
v_i = \frac{\sum_{j=1}^{n}q(k_i^j)}{\sum_{i=1}^{|\mathcal{K}|}\sum_{j=1}^{n}q(k_i^j)}.
\end{equation}

\par{Therefore, the proportion of skill sets in the exam paper $\bar{V}$ is defined as: $\bar{V}=[v_1,v_2,...,v_{|\mathcal{K}|}]$. Considering that we can not acquire the course skill weights, the skill occurrence probabilities of the questions in $\mathcal{Q}$ are used as the skill weights of the course $(\bar{C})$.}

\par{\textbf{Reward Function.} Finally, a sophisticated reward function is designed to simultaneously optimize three exam domain-specific objectives. It is defined as:}
\begin{equation}
r= \omega_1 r_1 + \omega_2 r_2 + \omega_3 r_3 , \{\omega_1, \omega_2, \omega_3\}\in [0,1] \label{e.4.13}
\end{equation}
where $\omega_1$, $\omega_2$, $\omega_3$ represents balance coefficients. 
\section{Experiment}\label{sec:5}
Our experimental study aims at addressing the following research questions. \textbf{(RQ1)} How does the proposed MOEPG performs compared with the state-of-the-art EPG methods? \textbf{(RQ2)} How does MOEPG’s performance varies with removing the question set partition mechanism? \textbf{(RQ3)} Can MOEPG effectively balance the proposed multiple objectives? \textbf{(RQ4)} Can MOEPG withstand parallel exam paper generation scenarios? 
\subsection{Experimental Setup.}
\par \
\par
\begin{figure}[htbp]
\begin{minipage}{0.49\linewidth}
			\centering
			\includegraphics[width=\linewidth]{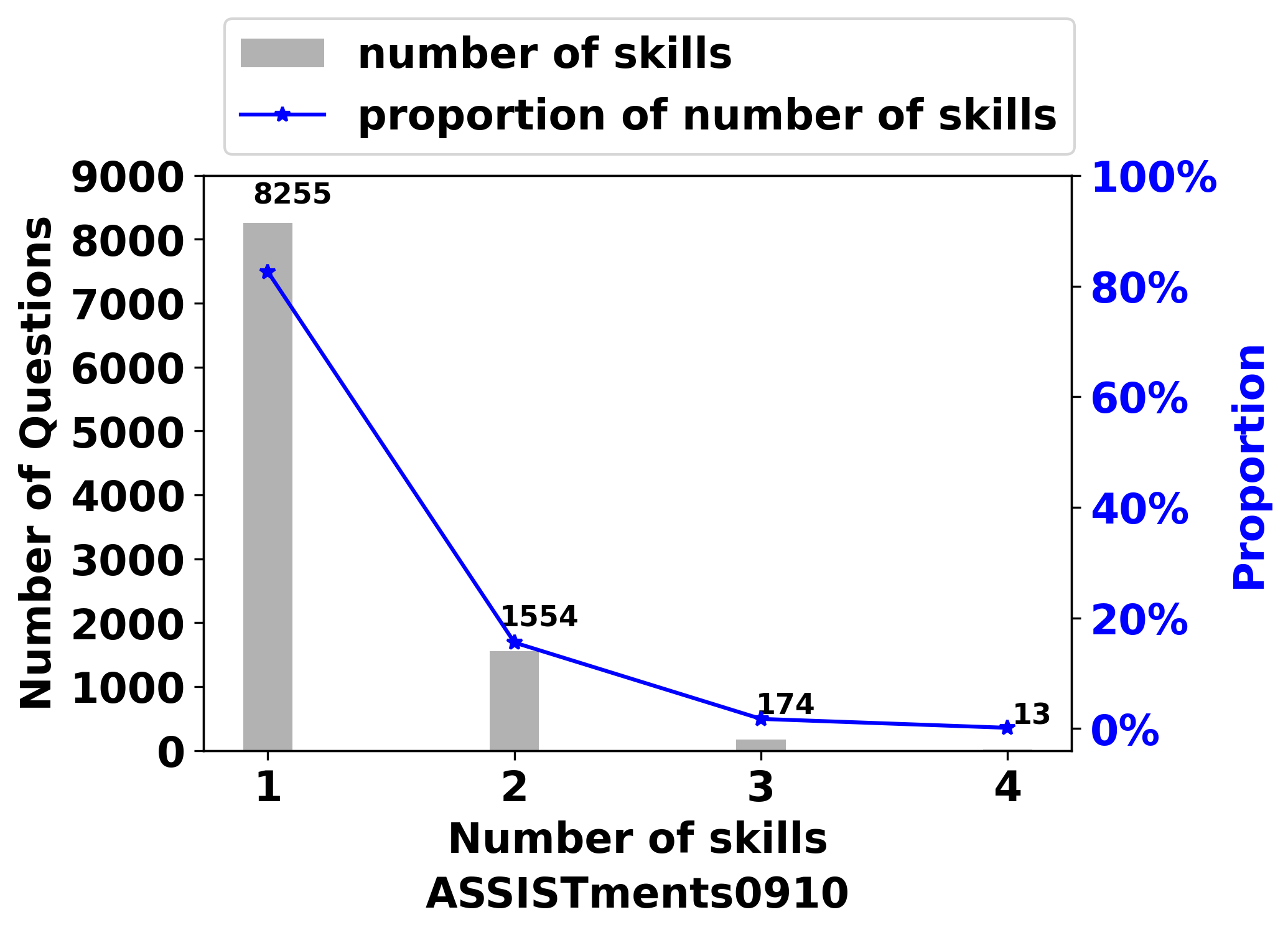}
		\end{minipage}
		\begin{minipage}{0.49\linewidth}
			\centering
			\includegraphics[width=\linewidth]{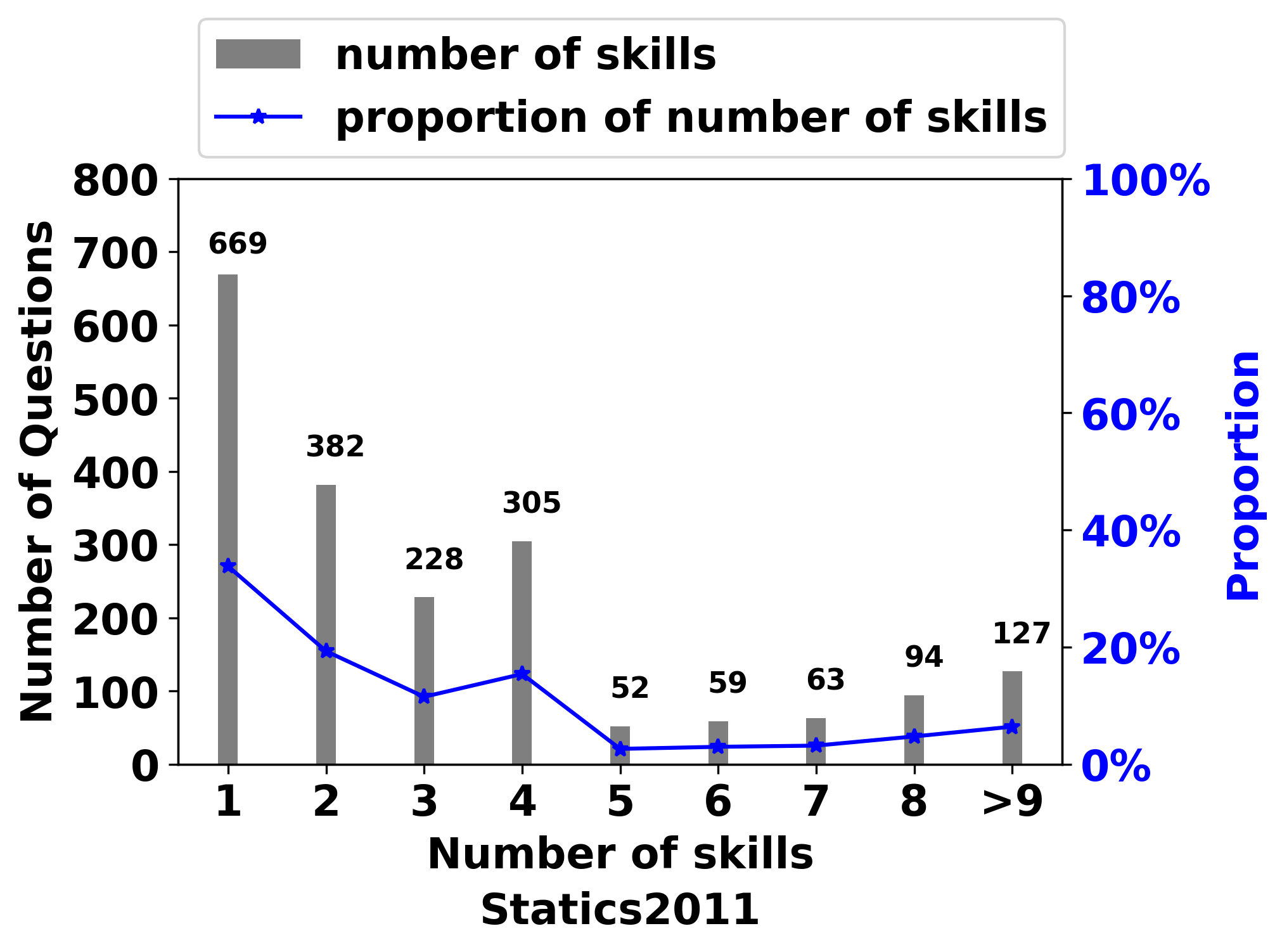}
		\end{minipage}
\caption{Statistics of similarities between questions.}\label{fig3}
\end{figure}

{\noindent \textbf{Datasets Description.} Two real-world datasets have been used to evaluate the effectiveness of MOEPG. We conduct detailed data analyses in Table \ref{tab1} and Figure \ref{fig3}. We observe that the number of questions in the ASSISTments0910 dataset is large, but the types are relatively poor; the Statics2011 dataset has a small number of questions, but the types are relatively rich.}

\par{\textbf{ASSISTments0910}\footnote{\href{https://sites.google.com/site/assistmentsdata/home/assistment-2009–2010-data}{https://sites.google.com/site/assistmentsdata/home/assistment-2009–2010-data}} is provided by ASSISTments online tutoring systems. The data is gathered from skill builder question sets. Examinees with no skills or less than three records were removed in preprocessing \cite{bib28}.}
\par{
\textbf{Statics2011}\footnote{\href{https://pslcdatashop.web.cmu.edu/DatasetInfo?datasetId=507}{https://pslcdatashop.web.cmu.edu/DatasetInfo?datasetId=507}}. is a dataset containing college student interactions on a one-semester engineering statics course. In our experiments, a concatenation of question name and step name is used as a knowledge skill.}
\begin{table}[!ht]
	\centering
	\caption{Detailed statistics of two real-world datasets.}
	\label{tab1}
	\resizebox{\linewidth}{!}{
	\begin{tabular}{ccccc}                &\\ 
		\cline{1-5}
		 & Statistics    & ASSISTments0910 & Statics2011 & \\ 
		\cline{1-5}
	
	 &	\#Skills 		 & 110 & 1,223           & \\
	 &	\#Examinees  & 4,151     & 333        &   \\
   & \#Questions  & 16,891     & 300       & \\
	 &	\#Records  & 325,637      & 189,287     &  \\
		\hline
	\end{tabular}
	}
\end{table}
\par{In order to generate the mock question set for the EPG task, we randomly selected 10,000 questions from the ASSISTments0910 dataset. For the statics2011 dataset, we refer to the skill distribution of 300 questions in the original dataset and randomly generated 1,700 new questions. By doing so, a synthetic question set containing 2,000 questions was formed, increasing the number of questions available for the MOEPG model. For the baseline approach, we define the difficulty label of each question as the question $q^{(j)}$’s right rate calculated from the historical learning interaction:}
\begin{equation}
difficulty_{q^{(j)}} = \frac{\sum_e^{|E|}\sum_x^{|\mathcal{X}|}y_j}{\sum_e^{|E|}\sum_x^{|\mathcal{X}|}q^{(j)}} ,\label{e.5.14}
\end{equation}
\noindent where $q^{(j)}$ represents whether the $j$-th question appears or not in the historical learning records.
\par{\noindent \textbf{Competitors.} We compare our model with various state-of-the-art baselines, including a representative method (RSF \cite{bib11}), one single-objective method (MCPSO \cite{bib7}), four many-objective methods (BACSTG \cite{bib2}, PGA-EG \cite{bib16}, MMGA \cite{bib15}, and MMGASA \cite{bib15}). }
\par{(1) \textbf{RSF}\cite{bib11} randomly selects questions from the question set to form the exam paper. We reserve the best version of the exam paper for comparison. }

\par{(2) \textbf{MCPSO}\cite{bib7} adopts a particle swarm optimization algorithm to generate multiple-choice tests. }

\par{(3) \textbf{BACSTG}\cite{bib2} generates exam paper by optimizing an objective function which is defined based on multiple objectives (i.e., discrimination degree, difficulty degree, and question type).}

\par{(4) \textbf{PGA-EG}\cite{bib16} regards a question as a chromosome that constitutes an exam paper and integrates difficulty and coverage as evolutionary objectives. }

\par{(5) \textbf{MMGA}\cite{bib15} designs a parallel migration genetic algorithm, which can simultaneously optimize the exam duration, number of questions, and difficulty level.}

\par{(6) \textbf{MMGASA}\cite{bib15} mixes strengths of the MGASA model as well as the simulated annealing algorithm.}

\begin{table*}
\centering
\caption{Overall performance (The best baselines are marked with asterisks, and the best results are boldfaced).}
\label{tab2}
\resizebox{\linewidth}{!}{
  \begin{tabular}{c|cc|cc|cc|ccc} 
  
		\cline{1-10}  
		
	\multirow{2}{*}{Model}	& \multicolumn{2}{c}{Difficulty}	\vline &  \multicolumn{2}{c}{Rationality} \vline & \multicolumn{2}{c}{Validity} &\multicolumn{2}{c}{Avg} & \\ 
		\cline{2-10} 
 	 & ASSISTments0910 & Statics2011 &	ASSISTments0910 & Statics2011 & ASSISTments0910 &	Statics2011 & ASSISTments0910 &	Statics2011 & 	 \\
	\cline{1-10} 
	RSF\cite{bib11} & 0.8826 $\pm$ 0.0142 	&0.9389 $\pm$ 0.0251&	0.8959	$\pm$ 0.0116& 0.8364 $\pm$ 0.0018 &0.8619 $\pm$ 0.0170	&0.6829 $\pm$ 0.0138	& 0.8801 $\pm$ 0.0143& 0.8194 $\pm$ 0.0136 & \\ 
	
	MCPSO\cite{bib7} & 0.8972 $\pm$ 0.0090 	&0.9543 $\pm$ 0.0134&	 0.9122	$\pm$ 0.0116& 0.8575 $\pm$ 0.0010 &0.8724 $\pm$ 0.0125	&0.6940 $\pm$ 0.0117 & 0.8939 $\pm$ 0.0110 & 0.8353 $\pm$ 0.0087
&\\
	
	BACSTG\cite{bib2} & 0.9024 $\pm$ 0.0089 	&0.9551 $\pm$ 0.0109&	0.9125	$\pm$ 0.0092& 0.8666 $\pm$ 0.0017 &0.8911 $\pm$ 0.0101	&0.6980 $\pm$ 0.0096	& 0.9020 $\pm$ 0.0094& 0.8399 $\pm$ 0.0074 &\\
	
	PGA-EG\cite{bib16} & 0.9199 $\pm$ 0.0072	&0.9712 $\pm$ 0.0067&	0.9257	$\pm$ 0.0074& 0.8953 $\pm$ 0.0008 &0.9030 $\pm$ 0.0090	&0.7199 $\pm$ 0.0070$^*$	& 0.9162 $\pm$ 0.0079& 0.8621 $\pm$ 0.0048$^*$ &\\
	
	MMGA\cite{bib15} & 0.9169 $\pm$ 0.0086	&0.9676 $\pm$ 0.0105&	0.9205	$\pm$ 0.0085& 0.8849 $\pm$ 0.0016 & 0.8983 $\pm$ 0.0104	&0.7051 $\pm$ 0.0096	& 0.9119 $\pm$ 0.0092
 & 0.8525 $\pm$ 0.0072&\\
 
	
	MMGASA\cite{bib15} & 0.9214 $\pm$ 0.0063	&0.9725 $\pm$ 0.0055 &	0.9306 $\pm$ 0.0063 & 0.8918 $\pm$ 0.0013 &0.9065 $\pm$ 0.0088	&0.7172 $\pm$ 0.0071	& 0.9195 $\pm$ 0.0071$^*$ & 0.8605 $\pm$ 0.0046&\\
	
	\cline{1-10}

	\emph{MOEPG-r1} & \textbf{0.9584 $\pm$ 0.0051}	& \textbf{0.9902 $\pm$ 0.0034} &	0.9603	$\pm$ 0.0062$^*$& 0.9065 $\pm$ 0.0031 & 0.4898 $\pm$ 0.0091	&0.3278 $\pm$ 0.0139	& 0.8028 $\pm$ 0.0068 &0.7415 $\pm$ 0.0068&\\
	

	\emph{MOEPG-r2} & 0.9319 $\pm$ 0.0038$^*$	& 0.9852 $\pm$ 0.0047 &	\textbf{0.9713	$\pm$ 0.0039} & \textbf{0.9232 $\pm$ 0.0010} & 0.4219 $\pm$ 0.0112	& 0.2761 $\pm$ 0.0125	& 0.7750 $\pm$ 0.0063 & 0.7282 $\pm$ 0.0061&\\
	

	\emph{MOEPG-r3} & 0.7899 $\pm$ 0.0081	& 0.4790 $\pm$ 0.0051 &	 0.8033	$\pm$ 0.0093& 0.4189 $\pm$ 0.0029 & \textbf{0.9850 $\pm$ 0.0053}	& \textbf{0.8895 $\pm$ 0.0046}	& 0.8594 $\pm$ 0.0076 &0.5958 $\pm$ 0.0042&\\
	
	\textbf{MOEPG} & 0.9315 $\pm$ 0.0042	& 0.9889 $\pm$ 0.0049&	0.9508	$\pm$ 0.0041 & 0.9099 $\pm$ 0.0007$^*$ & 0.9133 $\pm$ 0.0067$^*$	&0.7349 $\pm$ 0.0054$^*$	& \textbf{0.9319 $\pm$ 0.0050} &\textbf{0.8779 $\pm$ 0.0037}&\\
	
	\cline{1-10}

	\textbf{$p$-value} & 9.7237E-7	& 7.3543E-12&	3.2049E-14& 1.1673E-11 &0.0115	&6.8854E-9	& 1.6295E-17 &5.5563E-21&\\
\cline{1-10}
	
\end{tabular} 

}
\end{table*}

\par{\noindent \textbf{Implementation Details.} We implement the experiments using PyTorch on RTX 3080 GPU devices. We randomly select 50 examinees from each dataset to form a class, and the DKT model is trained to evaluate their skill mastery level of them. For all the experiments, each exam paper was generated using different randomly selected seeds. For each dataset, the number of questions to be generated in an exam $n=100$, and the score of each question is set to one point. $\omega_1$, $\omega_2$ and $\omega_3$ are set to one-third respectively. To ensure fairness, we generate 20 sets of exam paper with seven methods respectively. In MOEPG, the question embedding size $d_q=30$, the exam paper status dimension $d_s = n \times d_q$, the hidden vector dimension $d_h=200$, the batch size is set to 128, the number of subsets $f$ is set to 10 and the threshold $ts$ is set to 0.91/0.72 at the ASSISTments0910 and Staics2010 dataset, respectively. At each training epoch, the greedy parameter $\varepsilon$ is linearly decreased from 0.99 to 0.1 per training step. We set the replay memory size $me=2000$ and the discount rate $\beta=0.9$. We used the Adam algorithm for adjustment during training.}


\par{\noindent \textbf{Evaluation Indicators.} Following previous works \cite{bib15, bib16,bib29,bib30}, we employ three indicators to optimize the three objectives of exam paper generation.}
\par{
\textbf{Difficulty }is a widely used indicator for measuring the difficulty degree of the generated exam paper. }
\begin{equation}
Difficulty = 1-\frac{|Average_{score}-70|}{100} ,\label{e.5.15}
\end{equation} 
\par{\noindent where $Average_{score}$ indicates the average exam scores of the student group. Prior researches \cite{bib15, bib30} have shown that it is most reasonable to control the difficulty degree of the exam paper around 0.7. Specifically, if $Difficulty$ is higher than 0.75, it means that the exam paper is pretty easy. While $Difficulty$ is less than 0.45, it means that the exam paper is pretty difficult.}

\par{\textbf{Rationality }reflects the difference between the real exam score distribution $R_s$ and desirable exam score distribution $Z$. Just like the literature \cite{bib16, bib29, bib30} points out, the desirable exam score distribution is $Z \sim N(70, 15^2)$. Then, we define Rationality in Eq.(\ref{e.5.16})}
\begin{equation}
Rationality = 1- Div_{Wasserstein}(P(R_s), P(Z)) ,\label{e.5.16}
\end{equation}

\par{\noindent where $Div_{Wasserstein}(\cdot)$ is used to measure the similarity among $P(R_s)$ and $P(Z)$.}

\par{\textbf{Validity} reflects the similarity among the skill proportion of the exam paper $V^W$ and the skill weights of the course $C^W$.}
\begin{equation}\label{e.5.17}
Validity = 1- Cosine(V^W, C^W) .
\end{equation}
\subsection{ Overall Performance Comparison (For RQ1).}\label{sec:5.2}
\par{Table \ref{tab2} reports the mean values and standard deviations over the three indicators for each method. There are several observations: (1) In the single-objective case, the variants of MOEPG perform the best, followed by the MOEPG. Overall, all three different objectives can benefit exam paper generation and MOEPG method can find optimal exam paper by considering them simultaneously. (2) At a glance, MOEPG consistently outperforms all baseline methods on both datasets, which provides strong evidence that our MOEPG can well resist the conflict between multi-objectives. (3) In all cases, many-objective methods perform better than single-objective methods. (4) The column ‘Avg’ indicates that the average results of MOEPG across all evaluation indicators exceed the baseline. The reason is that the MOEPG has a comprehensive domain objective and a performance-guaranteed optimization method while other models do not. Comparisons of mean values and best values all demonstrate the superiority of MOEPG. Besides, to evaluate whether there is any statistical difference between MOEPG and the baseline methods, we conduct the $t$-test with a significant factor of 0.05 for both datasets. Due to space limitations, Table \ref{tab2} only reports the $t$-test result of MOEPG and the best baseline.}
\begin{figure}[htbp]
	\centering
\subfigure[Comparison of three indicators.]{
\label{fig4:1}
			\includegraphics[width=0.49\linewidth]{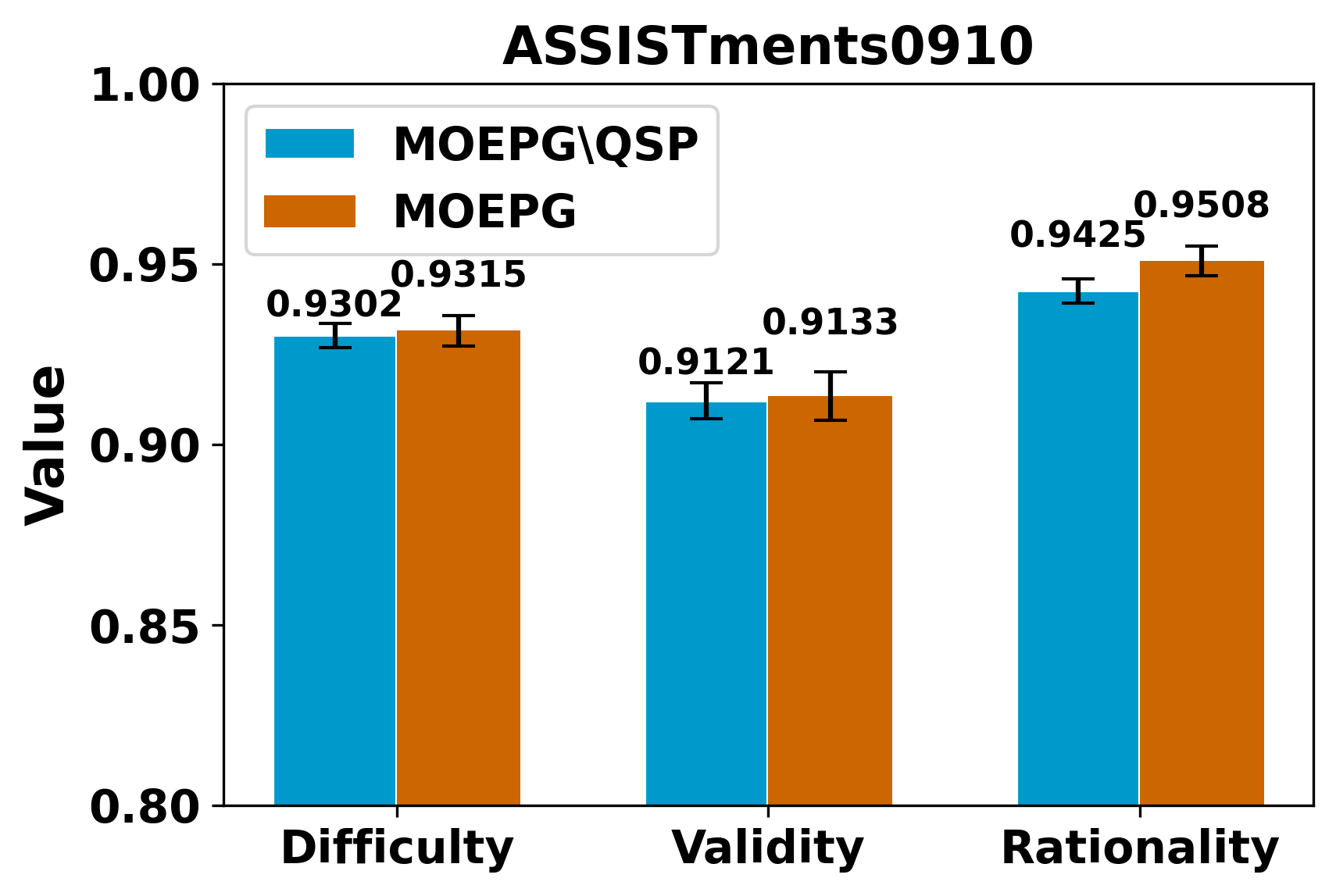}
			\includegraphics[width=0.49\linewidth]{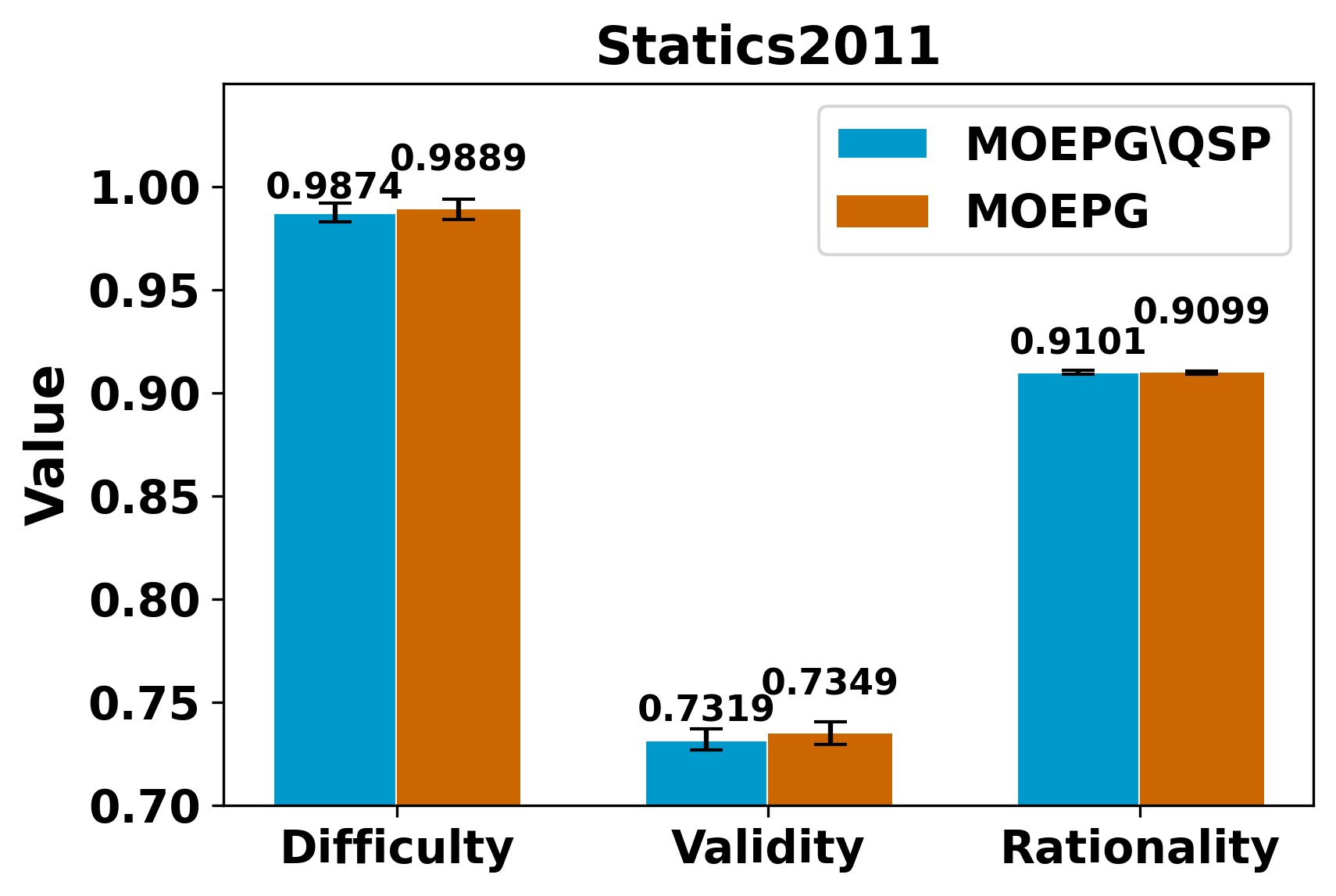}
}
\subfigure[Comparison of convergence curves.]{
\label{fig4:2}
			\includegraphics[width=0.49\linewidth]{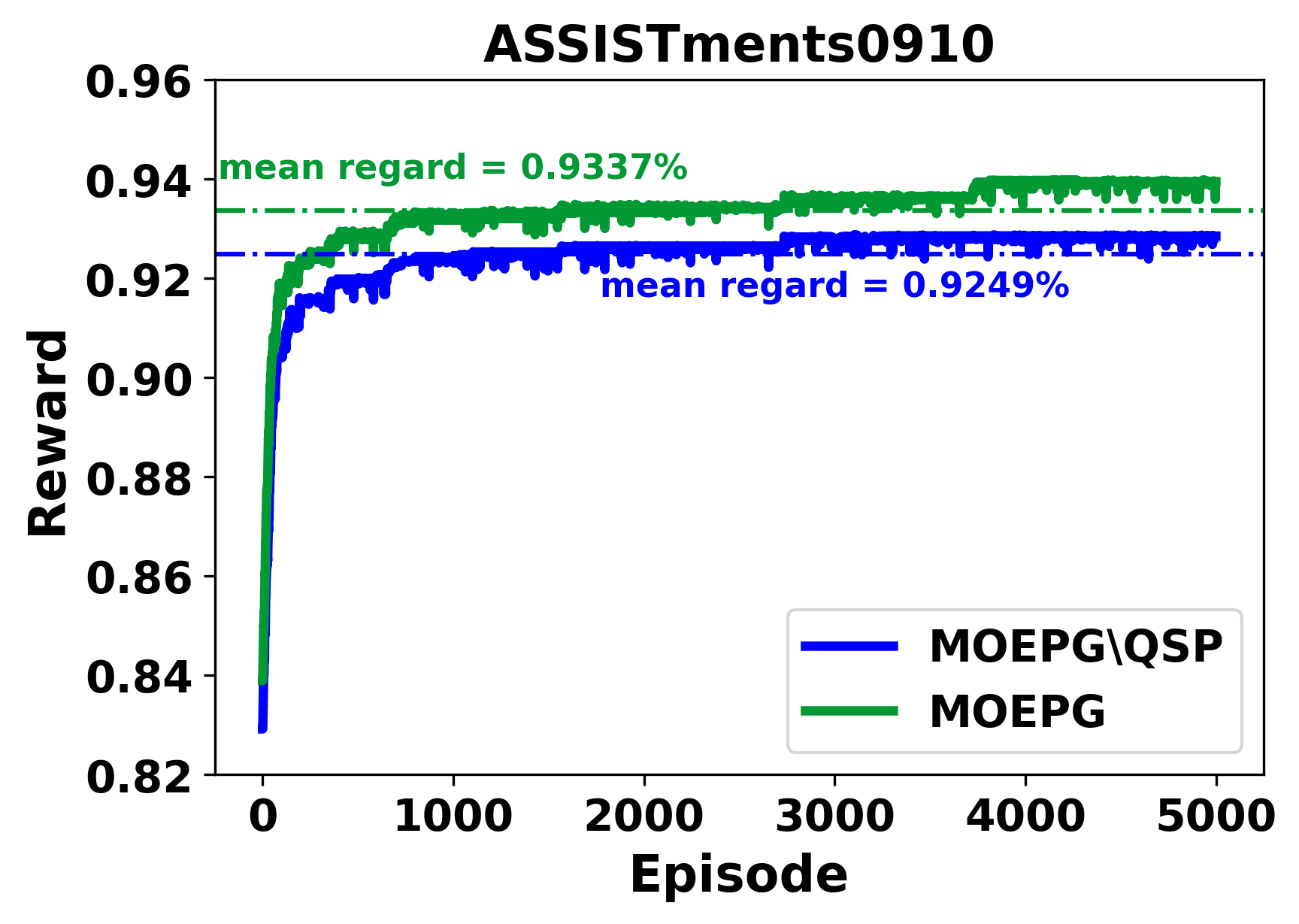}
			\includegraphics[width=0.49\linewidth]{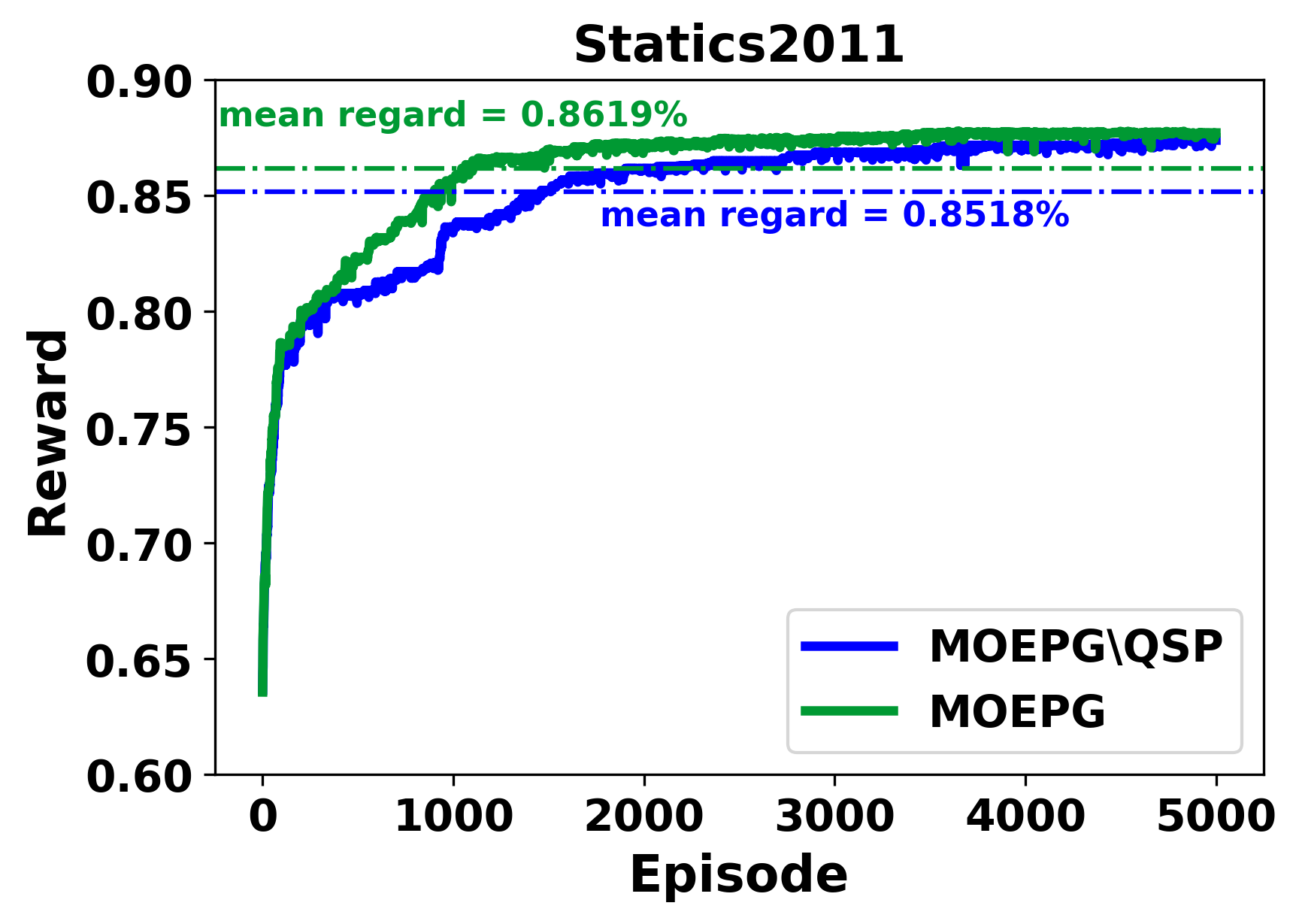}
}\caption{Ablation experimental results.}
\end{figure}
\vspace{-0.2cm}
\subsection{Evaluation on Question Set Partition (For RQ2).}\label{sec:5.3}
\par{We conduct the ablation study to verify the efficacy of the question set partition component. The variant MOEPG$\setminus $QSP removes this component, and thus adopts the global sampling strategy to update the exam paper. From Figure \ref{fig4:1} we can see that MOEPG$\setminus $QSP performs worse than MOEPG. Figure \ref{fig4:2} displays the cumulative reward convergence performance of 5,000 episodes attained by our default method MOEPG and its variant MOEPG$\setminus $QSP, where the x-axis represents the number of training episodes, and the y-axis represents the cumulative reward of each episode. The above observation clearly demonstrates the contribution of the question set partition component, and with full MOEPG outperforms MOEPG$\setminus $QSP.}

\vspace{-0.2cm}
\subsection{Evaluation of Pairwise Indicators (For RQ3).}\label{sec:5.4}
\par
\begin{figure}[htbp]
			\centering
			\includegraphics[width=\linewidth]{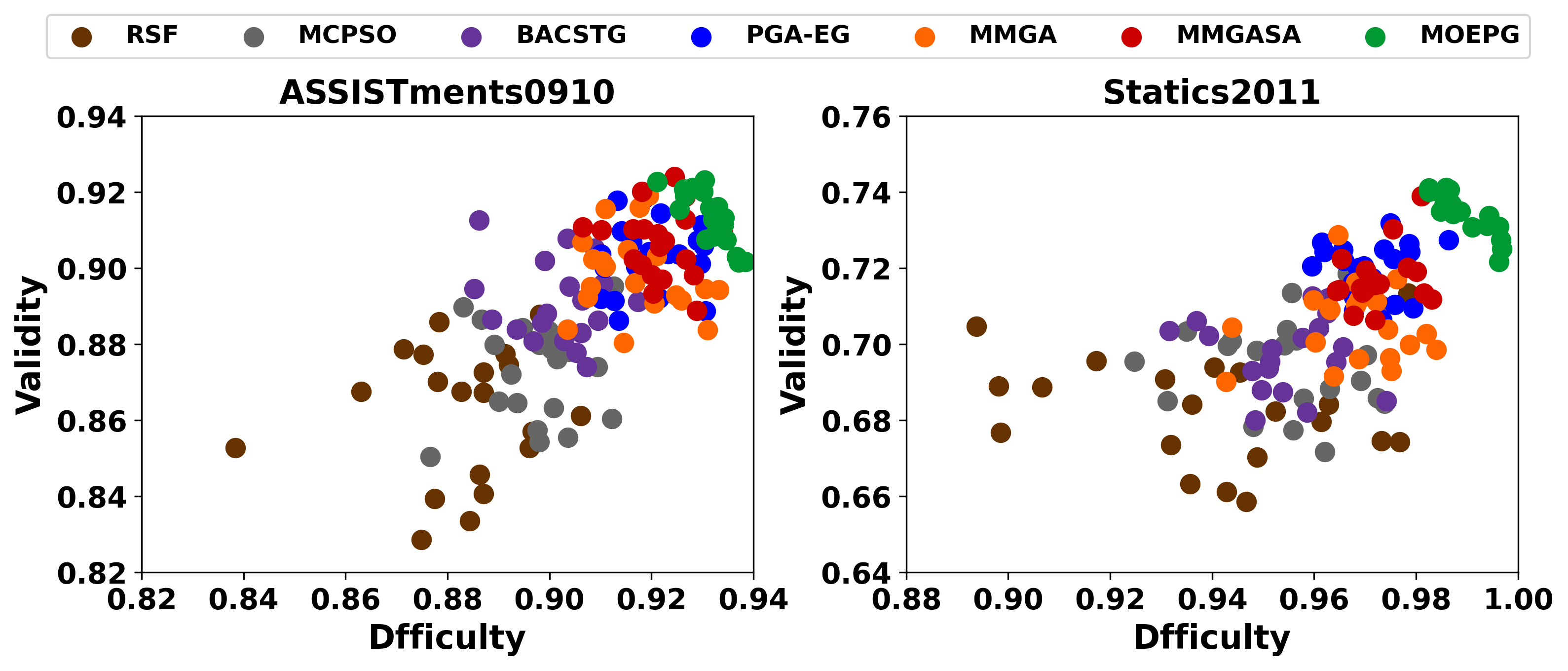}
\caption{Indicator comparison scatter plot.}\label{fig5}
\end{figure}

{To further show the superiority of our proposal, we conducted an experiment and analysis of the potential relationship between the two indicators. Figure \ref{fig5} depicts the scatter diagram of the quality of exam paper on two datasets. From the scatter plot, we see that the seven methods occupy slightly different clustering regions in the plot. The upper right corner of the scatter plot indicates the performance region where Difficulty and Validity achieve a fine balance. As indicated in Figure \ref{fig5}, MOEPG showed effectiveness in balancing Difficulty and Validity. Therefore, a comprehensive view shows that our MOEPG is closer to the ideal region.}
\vspace{-0.6cm}
\subsection{Model Scalability evaluation (For RQ4).}\label{sec:5.5}
\par{As online exams may involve many students, even if well-designed, they are prone to cheating. Therefore, it is necessary to generate $K$ similarly optimal exam papers, denoted as parallel exam paper ($K$-EPG) \cite{bib31}. We randomly selected three of the 20 exam papers generated by each method as parallel exam papers and evaluated them by estimating the duplication scale among the $K$ generated exams (see Eq.(\ref{e.5.18})). Please note that these 20 exam papers are consistent with the exam papers used for the assessment in section \ref{sec:5.2}. }
\begin{equation}\label{e.5.18}
Discrimination = 1-\frac{\sum_{i=1}^K(\sum_{j=1}^{K \setminus \{i\}}D_{ij})}{N},
\end{equation}
\par{\noindent where $N$ represents the total number of questions contained in the $K$ exam papers, and $D_{ij}$ is the number of duplication questions between exam $i$ and exam $j$.}

\begin{figure}[htbp]
\begin{minipage}{0.49\linewidth}
			\centering
			\includegraphics[width=\linewidth]{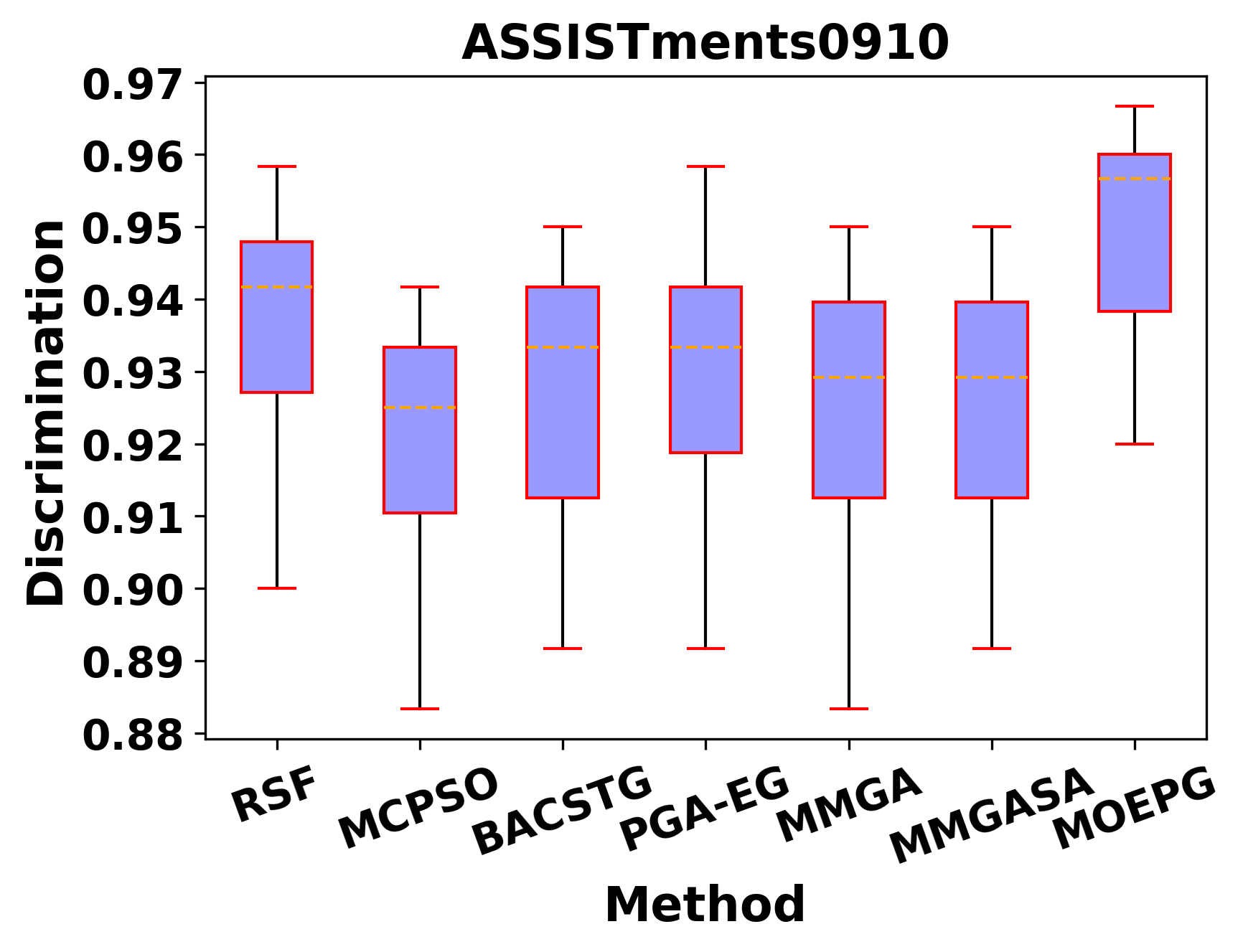}
		\end{minipage}
		\begin{minipage}{0.49\linewidth}
			\centering
			\includegraphics[width=\linewidth]{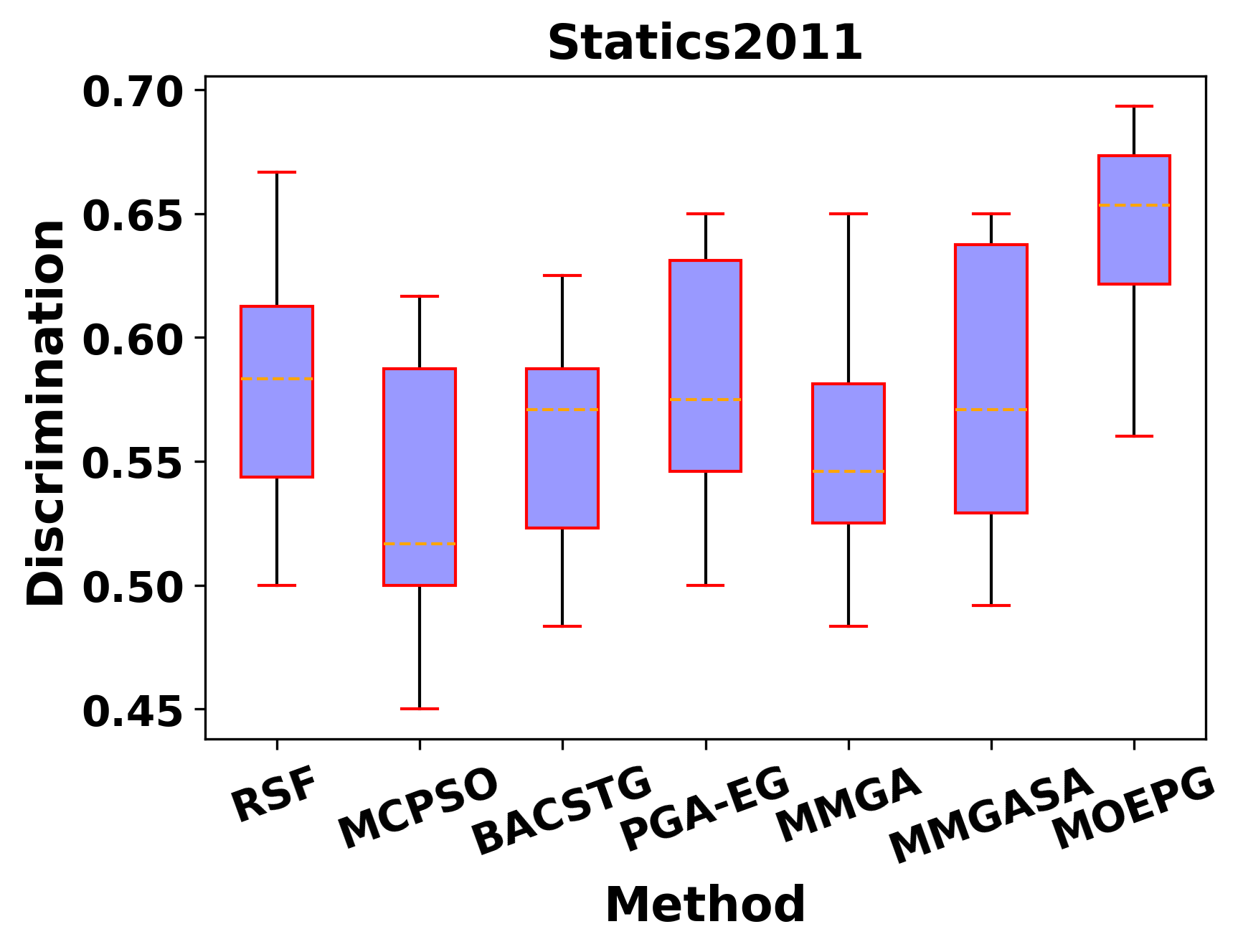}
		\end{minipage}
\caption{Performance results for K-EPG.}\label{fig6}
\end{figure}

\vspace{-0.3cm}
\par{Considering the stochastic nature of extraction process mean and standard deviation of results evaluated in 10 independent runs are used for the comparison. The results are displayed in Figure \ref{fig6}. We can conclude in two sides: (1) Our model outperforms all the baseline models, which demonstrates that MOEPG is promising in view of parallel exam paper generation scenarios. (2) Intuitively, MOEPG has comparable performance to the RSF methods. The main reason is that the RSF method focuses on boosting the randomness of the selection, which potentially reduces the occurrence of duplicate questions in the exam paper. (3) The best experiment result of Discrimination indicator generates from ASSISTments0910 dataset. A vital factor is that when the question set size and skill set size are imbalanced, there are far more questions involving the same skills, which makes MOEPG has more selectivity when generating parallel exam papers. Overall, the above observation confirms MOEPG can maintain effectiveness in different educational scenarios.}
\section{Conclusion}
\par{We are the first to integrate deep reinforcement learning into the exam paper generation domain, and thus using RL agent to generate the best suited exam paper for examinee group. Furthermore, our exam score prediction mechanism can acquire examinee’s knowledge mastery of multiple skills to further reflect the difficulty level of the exam paper, rather than merely relying on domain experts to label the difficulty level of the question. Later, the proposed MOEPG can filter some irrelevant questions and acquire candidate questions dynamically from a fresh perspective. Overall, the KT and the RL work together to outperform existing models and achieve start-of-the-art results. Furthermore, this work opens up plenty of opportunities for future research. First, our EPG framework can be extended from math courses to other courses (courses may vary in length, content, etc.). Second, cheating in examinations is an ongoing issue, so creating as many versions of the exam paper as there are examinees is another promising approach to ensure exam fairness.}

\section{Acknowledgement}
\par{This work was supported in part by the Educational Science Planning Project of Tianjin (Grant No. BIE210024), in part by the Science and Technology Program of Tianjin (Grant No. 22KPXMRC00210), in part by the National Key Research and Development Program of China (Grant No. 2021YFB1714800), in part by the Science and Technology Program of Tianjin (No. 22YDTPJC00940).}


\begin{thebibliography}{10}
\itemsep=0pt
\begin{small}
\bibitem{bib1} A. Ghosh, and A.Lan. BOBCAT: Bilevel Optimization-Based Computerized Adaptive Testing. In IJCAI, 2021.

\bibitem{bib2} M. L. Nguyen, et al. Large-scale multiobjective static test generation for web-based testing with integer programming. IEEE Transactions on Learning Technologies, 2012.

\bibitem{bib3} X. Cheng, et al. A Multi-Objective Optimization Approach for Question Routing in Community Question Answering Services. IEEE Transactions on Knowledge and Data Engineering, 2017.

\bibitem{bib4} K. Xiong, and X. Huang. Research on Auto-Generating Test Paper System Based on LDA and Genetic Algorithm. In ICSESS, 2018.

\bibitem{bib5} G. J. Hwang, et al. On the Development of a Computer-Assisted Testing System with Genetic Test Sheet-Generating Approach. IEEE Transactions on Systems Man Cybernetics-Systems, 2005.

\bibitem{bib6} M. {\.I}nce, et al. A novel hybrid fuzzy AHP-GA method for test sheet question selection. International Journal of Information Technology \& Decision Making, 2020.

\bibitem{bib7} T. Nguyen, et al. Multi-swarm single-objective particle swarm optimization to extract multiple-choice tests. Vietnam Journal of Computer Science, 2019.

\bibitem{bib8} T. Bui, et al. Application of Particle Swarm Optimization to Create Multiple-Choice Tests. Journal of Information Science $\&$ Engineering, 2018.

\bibitem{bib9} D. T. Phan, et al. StepDIRECT - A Derivative-Free Optimization Method for Stepwise Functions. In SIAM, 2022.

\bibitem{bib10} C. Piech, et al. Deep knowledge tracing. In NIPS, 2015.


\bibitem{p1} H. Peng, et al. Reinforced, incremental and cross-lingual event detection from social messages. IEEE Transactions on Pattern Analysis and Machine Intelligence, 2022.

\bibitem{bib11} M. S. R. Chim, et al. Automatic question paper generation using parametric randomization. J. Gujarat Res. Soc., 2019.

\bibitem{bib12} S. A. El-Rahman, et al. Automated test paper generation using utility based agent and shuffling algorithm. International Journal of Web-Based Learning and Teaching Technologies, 2019.

\bibitem{bib13} S. Kamya, et al. Fuzzy logic based intelligent question paper generator. In IACC, 2014.

\bibitem{bib14} T. N. T. Abd Rahim, et al. Automated exam question generator using genetic algorithm. In IC3e, 2017.


\bibitem{bib15} T. Nguyen, T.Bui, et al. Multiple-objective optimization applied in extracting multiple-choice tests. Engineering Applications of Artificial Intelligence, 2021.

\bibitem{bib16} Z. Wu, et al. Exam paper generation based on performance prediction of student group. Information Sciences, 2020.

\bibitem{bib17} P. Gu, et al. An Improved Personalized Genetic Algorithm Incorporated Item Distribution for Test Sheet Assembling. Applied Mathematics $\&$ Information Sciences, 2014.

\bibitem{bib19} A. T. Corbett, et al. Knowledge tracing: Modeling the acquisition of procedural knowledge. User Modeling and User-Adapted Interaction, 1994.

\bibitem{p2} H. Peng, et al. Reinforced neighborhood selection guided multi-relational graph neural networks. ACM Transactions on Information Systems, 2021.

\bibitem{bib20} P. Pavlik, et al. Performance Factors Analysis - A New Alternative to Knowledge Tracing. In FAIA, 2009.

\bibitem{bib21} Y. Zhuang, et al. Fully Adaptive Framework: Neural Computerized Adaptive Testing for Online Education. In AAAI, 2022.

\bibitem{bib22} P. Kantharaju, et al. Modeling Player Knowledge in a Parallel Programming Educational Game. IEEE Transactions on Games, 2022.

\bibitem{bib23} C. Wang, et al. Learning from Non-Assessed Resources: Deep Multi-Type Knowledge Tracing. In EDM, 2021.
 
\bibitem{bib24} H. V. Hasselt, et al. Deep reinforcement learning with double q-learning. In AAAI, 2016.

\bibitem{bib25} S. Zhou, et al. Interactive Recommender System via Knowledge Graph-enhanced Reinforcement Learning. In SIGIR, 2020.

\bibitem{bib26} H. Chen, et al. Large-Scale Interactive Recommendation with Tree-Structured Policy Gradient. In AAAI, 2019.

\bibitem{bib27} N. Tomasevic, et al. An overview and comparison of supervised data mining techniques for student exam performance prediction. Computers $\&$ Education, 2019.

\bibitem{bib28} S. Shen, et al. Convolutional Knowledge Tracing: Modeling Individualization in Student Learning Process. In SIGIR, 2020.

\bibitem{bib29} J. Leighton, et al. Cognitive diagnostic assessment for education: Theory and applications. Cambridge University Press, 2007.

\bibitem{p3} X. Zhao, et al. Multi-view tensor graph neural networks through reinforced aggregation. IEEE Transactions on Knowledge and Data Engineering, 2022.

\bibitem{bib30} W. Yuan, et al. The statistical analysis and evaluation of examination results of materials research methods course. Creative Education, 2013.

\bibitem{bib31} Y. Lin, et al. A discrete multiobjective particle swarm optimizer for automated assembly of parallel cognitive diagnosis tests. IEEE Transactions on Cybernetics, 2019.

\bibitem{bib32} X. Yang, et al. Rethinking rotated object detection with gaussian wasserstein distance loss. In ICML, 2021.

\bibitem{bib33} C. Haythornthwaite, et al. E-learning theory and practice. Sage Publications, 2011.

\end{small}
\end{thebibliography}
\end{document}